\pdfoutput=1

\documentclass[11pt, twocolumn]{article}

\usepackage[]{ACL2023}

\usepackage{times}
\usepackage{latexsym}
\usepackage{hyperref}
\usepackage{graphicx}
\usepackage{verbatim}
\usepackage{booktabs}
\usepackage{float}

\usepackage[T1]{fontenc}
\usepackage[utf8]{inputenc}
\usepackage{microtype}
\usepackage{inconsolata}

\newcommand{\name}[1][]{\textsc{CheXpert Plus}}

\usepackage{multicol} 

\usepackage{inconsolata}
\usepackage{listings}
\usepackage{color}
\usepackage{algorithm}
\usepackage{algpseudocode}
\usepackage{graphicx}
\usepackage{float}
\usepackage{pifont}
\usepackage{booktabs}
\usepackage{tabularx}
\usepackage{geometry}
\usepackage{amsmath}

\usepackage{times}
\usepackage{microtype}
\usepackage{epsfig}
\usepackage{caption}
\usepackage{float}
\usepackage{placeins}
\usepackage{color, colortbl}
\usepackage{stfloats}
\usepackage{enumitem}
\usepackage{tabularx}
\usepackage{xstring}
\usepackage{multirow}
\usepackage{xspace}
\usepackage{url}
\usepackage{subcaption}
\usepackage{xcolor}
\usepackage[hang,flushmargin]{footmisc}

\definecolor{mylightgray}{gray}{0.9}

\definecolor{codegreen}{rgb}{0,0.6,0}
\definecolor{codegray}{rgb}{0.5,0.5,0.5}
\definecolor{codepurple}{rgb}{0.58,0,0.82}
\definecolor{backcolour}{rgb}{0.95,0.95,0.92}
\definecolor{mypurple}{RGB}{200,192,248}
\definecolor{mypurpledeep}{RGB}{142,126,240}
\definecolor{mygreen}{RGB}{117,170,156}
\definecolor{myyellow}{RGB}{255,192,0}
\definecolor{myblue}{RGB}{57,143,255}
\definecolor{mygrey}{RGB}{231,230,230}
\definecolor{codey}{RGB}{220,220,170}
\definecolor{coder}{RGB}{206,145,120}
\definecolor{codeb}{RGB}{156,220,254}
\definecolor{codenum}{RGB}{204,204,204}

\newcommand{\cmark}{\textcolor{mygreen}{\ding{51}}} 
\DeclareSymbolFont{extraup}{U}{zavm}{m}{n}
\DeclareMathSymbol{\varheart}{\mathalpha}{extraup}{86}
\DeclareMathSymbol{\vardiamond}{\mathalpha}{extraup}{87}

\lstdefinestyle{mystyle}{
    backgroundcolor=\color{backcolour},   
    commentstyle=\color{codegreen},
    keywordstyle=\color{magenta},
    numberstyle=\tiny\color{codegray},
    stringstyle=\color{codepurple},
    basicstyle=\footnotesize,
    breakatwhitespace=false,         
    breaklines=true,                 
    captionpos=b,                    
    keepspaces=true,                 
    numbers=left,                    
    numbersep=5pt,                  
    showspaces=false,                
    showstringspaces=false,
    showtabs=false,                  
    tabsize=2
}

\title{CheXpert Plus: Augmenting a Large Chest X-ray Dataset with \\Text Radiology Reports, Patient Demographics and \\Additional Image Formats}

\author{
Pierre Chambon\thanks{\ \ Equal contribution}{$^{*\spadesuit}$} \\
  \texttt{pchambon@stanford.edu}
  \\\And
   Jean-Benoit Delbrouck$^{*\spadesuit{}}$ \\
  \texttt{jbdel@stanford.edu} \\
   \\\And
  Thomas Sounack$^\spadesuit$\\
  \texttt{tsounack@stanford.edu} \\
  \\\AND
   Shih-Cheng Huang$^\spadesuit$ \\
  \\\And
    Zhihong Chen$^\spadesuit$\\
  \\\And
    Maya Varma$^\spadesuit$\\
  \\\AND
  Steven QH Truong$^\heartsuit$\\
  \\\And
  Chu The Chuong$^\heartsuit$\\
  \\\And
   Curtis P. Langlotz$^\spadesuit$ \\\AND
   $^{\spadesuit}$ Stanford AIMI \hspace{0.2cm}   $^{\heartsuit}$VinBrain\
}

\begin{document}

\makeatletter
\let\@oldmaketitle\@maketitle
\renewcommand{\@maketitle}{\@oldmaketitle
  \centerline{\includegraphics[width=0.89\linewidth]{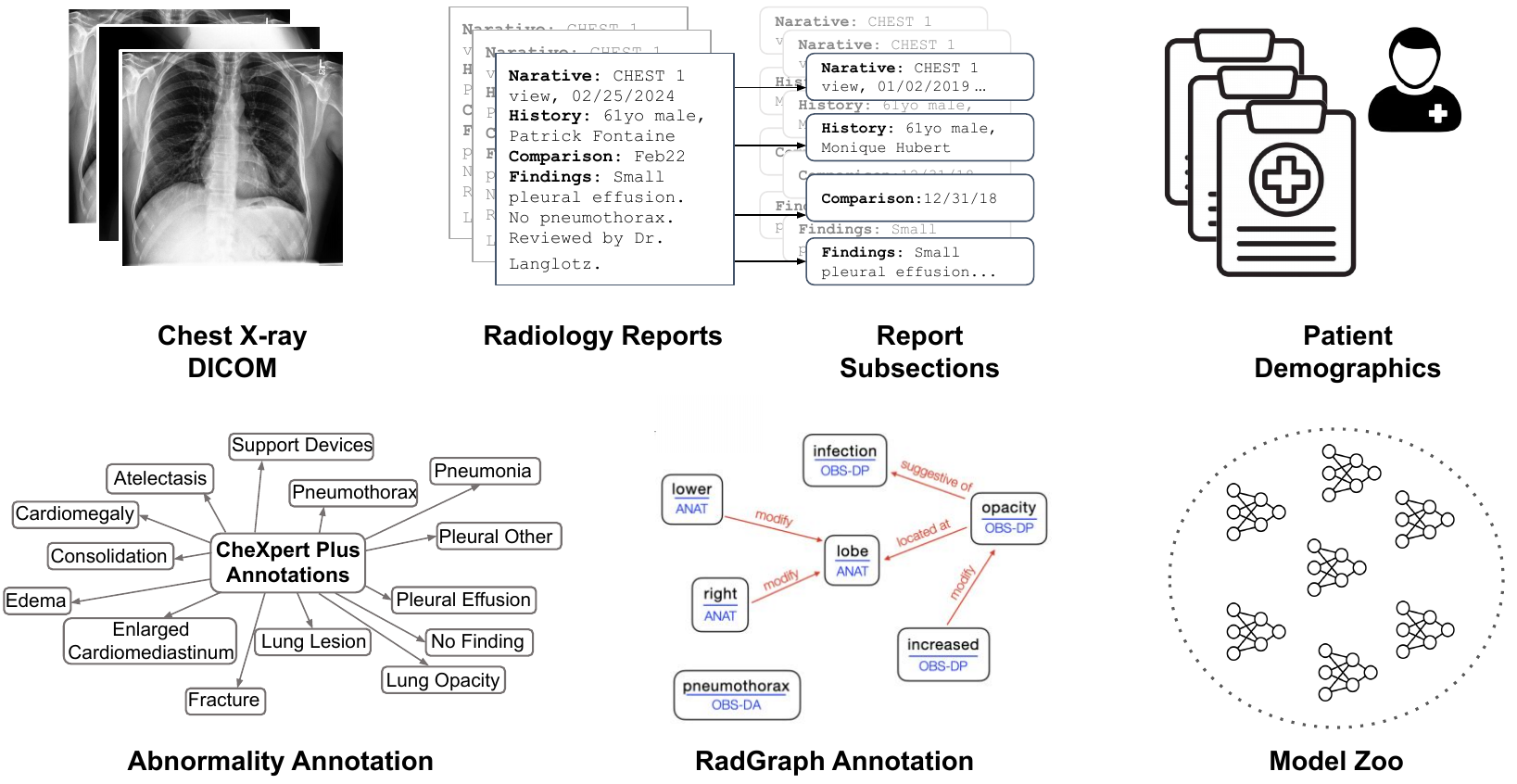}\bigskip}}
\makeatother
\maketitle

\begin{abstract}

Since the release of the original CheXpert paper \cite{irvin2019chexpert} five years ago, CheXpert has become one of the most widely used and cited clinical AI datasets.  
The emergence of vision language models has sparked an increase in demands for sharing reports linked to CheXpert images, along with a growing interest among AI fairness researchers in obtaining demographic data. To address this, \name~ serves as a new  collection of radiology data sources, made publicly available to enhance the scaling, performance, robustness, and fairness of models for all subsequent machine learning tasks in the field of radiology. \name~ is the largest text dataset publicly released in radiology, with a total of 36 million text tokens, including 13 million impression tokens. To the best of our knowledge, it represents the largest text de-identification effort in radiology, with almost 1 million PHI spans anonymized. It is only the second time that a large-scale English paired dataset has been released in radiology, thereby enabling, for the first time, cross-institution training at scale. All reports are paired with high-quality images in DICOM format, along with numerous image and patient metadata covering various clinical and socio-economic groups, as well as many pathology labels and RadGraph annotations. We hope this dataset will boost research for AI models that can further assist radiologists and help improve medical care. Data is available at the following URL:{\small~\url{https://stanfordaimi.azurewebsites.net/datasets/5158c524-d3ab-4e02-96e9-6ee9efc110a1}} Models are available at the following URL:{\small~\url{https://github.com/Stanford-AIMI/chexpert-plus}}
\end{abstract}

\section{Introduction}

The rapid advancement of deep learning technologies has catalyzed transformative changes across numerous fields, with healthcare standing out as a particularly promising domain for such innovations. Integration of artificial intelligence into the analysis of chest X-rays has emerged as a significant area of progress and study. Early work has yielded models that rival the diagnostic capabilities of radiologists (cite CheXpert PLOS). Subsequent research has leveraged self-supervised pre-training objectives that bridge radiology reports with chest X-ray images, achieving enhanced performance metrics with considerably fewer data requirements~\cite{huang2023self}. More recently, the development of Vision Language Models (VLMs) aimed at generating chest X-ray reports has begun to illustrate the vast potential of AI to improve diagnostic processes. These advancements collectively move us toward a future where AI significantly enhances patient outcomes through more accurate and efficient diagnoses.

In recent years, there has been a surge in research developing Vision Language Models (VLMs) for chest X-ray analysis. Some key developments in this area include generating radiology reports~\cite{chen2020generating, delbrouck2022improving, hyland2023maira, chaves2024training}, employing self-supervised learning techniques~\cite{huang2021gloria, zhang2022contrastive, varma2023villa, bannur2023learning}, using stable diffusion models~\cite{chambon2022roentgen}, and applying broad reasoning foundation models~\cite{moor2023med,wu2023towards,tu2024towards,chen2024chexagent}.

However, the success of these technologies underscores the critical need for extensive datasets that encompass both images and text to train increasingly sophisticated models. This paper presents a significant update to the CheXpert dataset, accompanied by the release of corresponding radiology reports and patient information. We have undertaken a rigorous de-identification process to ensure that these reports are devoid of any patient Personal Health Information (PHI), thereby aligning with the ethical standards required for the utilization of medical data in research. This enhancement of the CheXpert dataset marks a pivotal step forward in the development of AI tools capable of transforming patient care in radiology and beyond.

Our new dataset, referred to as \name, comprises the following sources of data:

\textbf{Images}, available in both DICOM and PNG format.

\textbf{Reports}, corresponding to each CheXpert image, parsed into subsections.

\textbf{Demographics}, including several clinical and socio-economic attributes.

\textbf{14 pathology labels}, automatically extracted from the reports.

\textbf{RadGraph annotations}, extracted separately from Findings and Impression sections.

\textbf{Models}, trained on these data sources for key machine learning tasks.
\section{Related Work}
\begin{table*}[ht]
\centering

\label{tab:dataset_comparison}
\begin{tabularx}{\textwidth}{l *{6}{>{\raggedright\arraybackslash}X}}
\toprule[1.5pt]
\textbf{Dataset} & \textbf{Patients} & \textbf{X-rays} & \textbf{Labels} & \textbf{Reports} & \textbf{DICOM} & \textbf{Meta} \\
\midrule[0.7pt]
MIMIC-CXR~\cite{johnson2019mimic}  & 65,379 & 377,095 & 14 & \cmark & \cmark & \cmark \\
OpenI~\cite{demner2012design}      & 3,996  & 8,121   & -  & \cmark &  -     & -     \\
CheXpert~\cite{irvin2019chexpert}  & 64,540 & 223,414 & 14 & -      & \cmark & - \\
BraX~\cite{reis2022brax}           & 18,442 & 40,967  & 14 & -      & -      & -     \\
CandidPTX~\cite{Feng2021}          & 13,744 & 19,234  & 3  & -      & \cmark & -     \\
NIH~\cite{wang2017chestx}          & 30,805 & 112,120 & 14 & -      & -      & -     \\
PadChest~\cite{bustos2020padchest} & 67,625 & 160,861 & -  & \cmark*& -      & \cmark \\
VinDR~\cite{nguyen2020vinbigdata}  & 15,000 & 15,000  & 14 & -      & \cmark & \cmark \\
MIDRC                              & 131,351& 131,351 & 1  & -      & -      & \cmark \\
JF Healthcare~\cite{jfhealthcare}  & 16,000 & 16,000  & 1  & -      & -      & -     \\
\midrule[0.7pt]
\cellcolor{mylightgray} \textbf{CheXpert Plus}                      &\cellcolor{mylightgray}  64,725      & \cellcolor{mylightgray} 223,462 & \cellcolor{mylightgray}  14 & \cellcolor{mylightgray} \cmark & \cellcolor{mylightgray} \cmark & \cellcolor{mylightgray} \cmark     \\
\bottomrule[1.5pt]
\end{tabularx}

\caption{Comparison of Chest X-ray Datasets. *PadChest reports are in Spanish}
\label{tab:related_dataset}
\end{table*}

Several datasets comprising chest X-rays and their corresponding reports have been published, with the main ones outlined in Table ~\ref{tab:related_dataset}. Notably, the PadChest dataset~\cite{bustos2020padchest}, created from chest X-rays collected at the Hospital Universitario de San Juan, Alicante, Spain, between January 2009 and December 2017, encompasses 109,931 studies and 168,861 images, all de-identified for research purposes. It includes radiology reports written in Spanish with comprehensive metadata.  The data has been processed for quality and consistency, excluding non-compliant images based on specific criteria such as readability, incorrect modality, or inappropriate projections. 

The MIMIC Chest X-ray Database~\cite{johnson2019mimic} offers a collection of 377,110 de-identified chest radiographs in DICOM format, paired with free-text radiology reports from 227,835 studies at Beth Israel Deaconess Medical Center (Boston, MA). The de-identification process included generating random identifiers for patients and studies while preserving chronological data integrity. A custom algorithm was developed to de-identify the chest radiographs while retaining medically relevant information. 

Other datasets, smaller in size, are also available to the research community, such as Open-I~\cite{demner2012design}, BIMCV-COVID19~\cite{vaya2020bimcv} and CANDID-PTX~\cite{feng2021curation}. 
\section{Dataset Composition}

\subsection{General Composition}
\label{sec:generalcomp}

\name~is a dataset that pairs text and images, featuring 223,228 unique pairs of radiology reports and chest X-rays from 187,711 studies and 64,725 patients. A single patient may be linked to several studies, and each study may include multiple chest X-rays. The dataset comprises:
\begin{itemize}
    \item 223,228 unique chest X-ray images in DICOM format, each featuring 47 DICOM metadata elements. These images are also be made available in PNG format(Section~\ref{sec:images}).
    \item 187,711 unique radiology reports, each report divided into subsections extracted from the original corresponding report (Section~\ref{sec:reports}).
    \item 64,725 unique patients, with 8 de-identified demographic data points (Section~\ref{sec:patients}).
    \item 187,711 unique annotations for 14 different chest pathologies using CheXbert.
    (Section~\ref{sec:chexpertlabels}).
    \item 187,575 unique RadGraph annotations for impression sections and 47,328 unique RadGraph annotations for finding sections, using the pretrained RadGraph model (Section~\ref{sec:radgraph}).
    \item a plurality of models trained on these data sources for key radiology tasks (Section~\ref{sec:modelzoo}).
\end{itemize}

\subsection{Images} 
\label{sec:images}

\name~comprises 223,228 unique images available both in DICOM format and in PNG format.

The DICOM format includes image metadata attributes that encapsulate essential information for medical image processing and interpretation. In total, we release up to 47 DICOM metadata elements, listed in Appendix~\ref{app:dicom_meta}. With these image metadata attributes, it is possible to convert the DICOM pixel data into the PNG format. While the DICOM format offers the most comprehensive data, the PNG format may be more straightforward to integrate into existing training pipelines.
\subsection{Reports} 
\label{sec:reports}

\begin{table*}[ht!]
\caption{Aggregated statistics for all reports and report sections as present in \name~. The BERT-base-uncased tokenizer is being used to compute report tokens.}
\label{table-reports-counts}
\centering
\begin{tabular}{l r r r r r r r} 
\toprule[1.5pt]
\cellcolor[HTML]{EFEFEF}Report Section
& \multicolumn{1}{c}{\cellcolor[HTML]{EFEFEF}\begin{tabular}[c]{@{}c@{}}Studies \\{Total}\end{tabular}}
& \multicolumn{1}{c}{\cellcolor[HTML]{EFEFEF}\begin{tabular}[c]{@{}c@{}}Tokens \\{Total}\end{tabular}}
& \multicolumn{1}{c}{\cellcolor[HTML]{EFEFEF}\begin{tabular}[c]{@{}c@{}}Tokens \\{Mean}\end{tabular}}
& \multicolumn{1}{c}{\cellcolor[HTML]{EFEFEF}\begin{tabular}[c]{@{}c@{}}Tokens \\{Std}\end{tabular}}
& \multicolumn{1}{c}{\cellcolor[HTML]{EFEFEF}\begin{tabular}[c]{@{}c@{}}Tokens \\{Median}\end{tabular}}
& \multicolumn{1}{c}{\cellcolor[HTML]{EFEFEF}\begin{tabular}[c]{@{}c@{}}Tokens \\{$\eta_{.1}$}\end{tabular}}
& \multicolumn{1}{c}{\cellcolor[HTML]{EFEFEF}\begin{tabular}[c]{@{}c@{}}Tokens \\{$\eta_{.9}$}\end{tabular}}
\\ 
\midrule
\addlinespace
Full Report
& 187,711
& 36,469,132
& 194
& 62
& 181
& 135
& 268
\\ 
Impression
& 187,575
& 13,351,758
& 71
& 33
& 66
& 34
& 113
\\ 
Findings
& 47,328
& 4,844,613
& 102
& 56
& 90
& 49
& 171
\\ 
Narrative
& 183,022
& 2,524,834
& 13
& 6
& 13
& 10
& 18
\\ 
Clinical History
& 116,231
& 1,661,044
& 14
& 6
& 14
& 7
& 22
\\ 
History
& 30,865
& 411,485
& 13
& 6
& 12
& 7
& 20
\\ 
Comparison
& 178,893
& 1,707,006
& 9
& 6
& 8
& 4
& 16
\\ 
Technique
& 9,173
& 105,376
& 11
& 7
& 10
& 8
& 14
\\ 
Procedure Comments
& 20,563
& 179,681
& 8
& 4
& 8
& 8
& 8
\\ 
End of Impression
& 9,033
& 216,481
& 23
& 32
& 4
& 2
& 65
\\ 
Summary
& 160,081
& 4,147,222
& 25
& 16
& 25
& 8
& 42
\\ 
\bottomrule[1.5pt]
\end{tabular}
\end{table*}

\name~contains 187,711 distinct radiology reports, each corresponding to a separate study. For studies comprising more than one image, the report compiles information extracted from all the images in the study. Therefore, the findings described in the report may describe findings from multiple images. Additionally, a single study may be preceded by one or more related studies, especially in cases where a patient has undergone multiple examinations. To facilitate the analysis of a disease's progression over multiple studies, we have introduced the field $patient\_report\_date\_order$, which ranks the studies of each patient in chronological order. This enhancement makes \name~compatible with ML approaches that condition their predictions not only on a single study but also on its prior studies. \\

Section names were not used consistently by the radiologists producing these reports.  Therefore, the reports from \name~may contain as many as 11 distinct sections:

\textbf{Narrative:} This section outlines the type of exam and the date it was conducted.

\textbf{Clinical History:} This details the reasons the ordering provider requested the radiology exam. It typically includes patient metadata and symptoms.

\textbf{History:} Similar to the Clinical History, this section explains the reasons for the radiology exam provided by the requesting provider, often incorporating patient symptoms and a reason for the exam.

\textbf{Comparison:} Here, any previous studies that were used as part of the interpretation are listed.

\textbf{Technique:} This section provides details on how the exam was conducted, including the views captured and whether contrast material was injected.

\textbf{Procedure Comments:} Noted here are additional details about the procedure, such as the number of views.

\textbf{Findings:} Observations reported by the radiologist are listed here in detail.

\textbf{Impression:} This section summarizes the findings and the radiologist's interpretation.

\textbf{End of Impression:} Closing remarks and often the reporting radiologist's name are included in this section.

\textbf{Summary:} A brief overview of the study is provided here, sometimes alongside a number that classifies the study as normal, abnormal, or extremely abnormal.

\textbf{Accession Number:} A de-identified accession number is given. Although this section holds limited informational value for any ML models, it is retained to preserve the original format of the report. \\

Aggregated statistics for each section are made available in Table~\ref{table-reports-counts}. In \name~reports, an impression is almost always present.

\subsection{Demographics}
\label{sec:patients}

\begin{figure}[t!]
\begin{center}
 \includegraphics[width=0.38\textwidth]{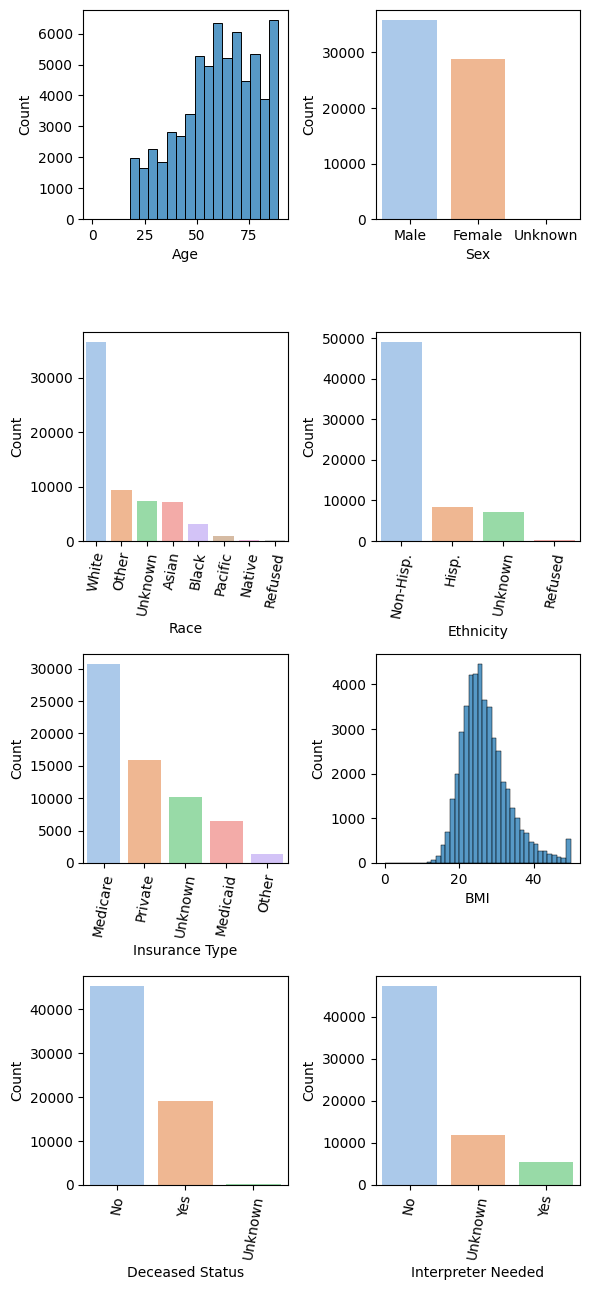}
\caption{Frequency plots for patient metadata as available in \name.}
\label{fig:csv_meta_data}
\end{center}
\end{figure}

Alongside the de-identified reports and images, \name~includes de-identified demographic data. Each study is linked to the corresponding demographic data using a privacy-preserving identifier created during the de-identification process. While a patient's insurance may have varied over time, the data contains their insurance status as of February 2024.

As listed in Figure \ref{fig:csv_meta_data}, \name~lists patient age, sex, race, ethnicity, insurance type, BMI, deceased status, and the need for an interpreter. Collectively, this demographic data allows researchers to account for subgroups of patients and improve the training of ML models relying on \name~data sources, enhancing their fairness and robustness.

\begin{table*}[h!]
\caption{14 pathology labels as present in the CheXpert Plus dataset. Total counts and proportions for each pathology are being reported, for the training set and counting each image of the dataset per CheXpert label}.
\label{table-disease-labels}
\centering
\resizebox{\linewidth}{!}{%
\begin{tabular}{lrrrr} 
\toprule[1.5pt]
\cellcolor[HTML]{EFEFEF}Pathology
& \cellcolor[HTML]{EFEFEF}Positive (\%)
& \cellcolor[HTML]{EFEFEF}Uncertain (\%)
& \cellcolor[HTML]{EFEFEF}Negative (\%)
& \cellcolor[HTML]{EFEFEF}Not Mentioned (\%)
\\ 
\midrule
Atelectasis
& 33,385 (14.94)
& 33,725 (15.09)
& 1,326 (0.59)
& 155,026 (69.37)
\\
Cardiomegaly
& 26,996 (12.08)
& 8,095 (3.62)
& 11,126 (4.98)
& 177,245 (79.32)
\\
Consolidation
& 14,790 (6.62)
& 27,727 (12.41)
& 28,116 (12.58)
& 152,829 (68.39)
\\
Edema
& 52,245 (23.38)
& 12,984 (5.81)
& 20,735 (9.28)
& 137,498 (61.53)
\\
Enlarged Cardiomediastinum
& 10,789 (4.83)
& 12,403 (5.55)
& 21,656 (9.69)
& 178,614 (79.93)
\\
Fracture
& 9,049 (4.05)
& 644 (0.29)
& 2,512 (1.12)
& 211,257 (94.54)
\\
Lung Lesion
& 9,193 (4.11)
& 1,486 (0.66)
& 1,271 (0.57)
& 211,512 (94.65)
\\
Lung Opacity
& 105,567 (47.24)
& 5,602 (2.51)
& 6,606 (2.96)
& 105,687 (47.30)
\\
No Finding
& 22,407 (10.03)
& 0 (0.00)
& 0 (0.00)
& 201,055 (89.97)
\\
Pleural Effusion
& 86,174 (38.56)
& 11,629 (5.20)
& 35,425 (15.85)
& 90,234 (40.38)
\\
Pleural Other
& 3,521 (1.58)
& 2,654 (1.19)
& 315 (0.14)
& 216,972 (97.10)
\\
Pneumonia
& 6,042 (2.70)
& 18,771 (8.40)
& 2,806 (1.26)
& 195,843 (87.64)
\\
Pneumothorax
& 19,453 (8.71)
& 3,141 (1.41)
& 56,347 (25.22)
& 144,521 (64.67)
\\
Support Devices
& 116,004 (51.91)
& 1,079 (0.48)
& 6,136 (2.75)
& 100,243 (44.86)
\\
\bottomrule[1.5pt]
\end{tabular}
}
\end{table*}

\subsection{Pathology Labels}
\label{sec:chexpertlabels}

\name~lists 14 pathology labels, generated from the radiology reports using model-based extraction methods. Table \ref{table-disease-labels} presents the prevalence of each pathology based on the CheXpert labeler \cite{irvin2019chexpert}. For each disease, the label can be positive (1), uncertain (-1), negative (0) or not mentioned.

In addition to these labels, \name~also includes labels generated from CheXbert \cite{smit2020chexbert}. Due to CheXbert's maximum input token size of 512 tokens, different parts of the report are used as input: the full report, the findings, the impression, as well as concatenations (findings - impression) and (impression - findings). When an input exceeds 512 tokens, the excess tokens are not used by the labeler.

To assess the performance of each label, these labels are compared against two human-annotated test sets. The first test set comprises 1000 samples annotated by 2 board-certified radiologists based on the radiology images only, with disagreement resolution through consensus. The second test set includes 500 samples annotated by 8 board-certified radiologists based on the radiology reports only, with a majority vote of 5 radiologists. In order to have meaningful comparisons, the test sets are restricted to instances where both findings and impression are available, leading to 154 and 339 labeled samples respectively. In order to simplify this multi-label classification problem, Not Mentioned labels are assigned to Negative and Uncertain labels to Positive. The results of this analysis are detailed in table \ref{tab:chexbert_labels_comparison}.

\begin{table}[!h]
\centering
\caption{Comparison of labelers on human-annotated sets, based on image and text respectively. The metrics reported are macro-averaged across the 14 pathology labels.}
\label{tab:chexbert_labels_comparison}
\resizebox{0.5\textwidth}{!}{
\begin{tabular}{@{}lcc@{}}
\toprule[1.5pt]
 & \multicolumn{1}{c}{Image Test \textit{(S. 154)}} &  \multicolumn{1}{c}{Text Test \textit{(S. 339)}} \\
\cmidrule(lr){2-2} \cmidrule(lr){3-3}
\cellcolor[HTML]{EFEFEF}Labeler & \multicolumn{1}{c}{\cellcolor[HTML]{EFEFEF}F1 (Pr., R.)} & \multicolumn{1}{c}{\cellcolor[HTML]{EFEFEF}F1 (Pr., R.)}\\
\midrule
CheXpert & 0.35 (0.40, 0.52) & 0.92 (0.91, 0.95)\\
\midrule
\begin{tabular}{@{}l@{}}CheXbert\\\textit{Full Report}\end{tabular} & 0.41 (0.41, 0.64) & 0.74 (0.67, 0.90)\\
\midrule
\begin{tabular}{@{}l@{}}CheXbert\\\textit{Findings}\end{tabular} & \textbf{0.44} (0.44, 0.59) & 0.65 (0.65, 0.72) \\
\midrule
\begin{tabular}{@{}l@{}}CheXbert\\\textit{Impression}\end{tabular} & 0.34 (0.39, 0.50) & \textbf{0.93} (0.92, 0.94)\\
\midrule
\begin{tabular}{@{}l@{}}CheXbert\\\textit{Findings + Impression}\end{tabular} & 0.42 (0.41, 0.64) & 0.77 (0.70, 0.93) \\
\midrule
\begin{tabular}{@{}l@{}}CheXbert\\\textit{Impression + Findings}\end{tabular} & 0.42 (0.41, 0.64) & 0.77 (0.69, 0.93)\\
\bottomrule[1.5pt]
\end{tabular}
}
\end{table}

For prediction tasks based on the reports, we recommend using the CheXbert-impression labels, which achieve the highest F1-score (0.93) when evaluated against the text-based test set. On the contrary, for prediction tasks based on the images, we suggest using the CheXbert-findings labels, which achieve an F1-score of 0.44 on the image-based test set. The performance gap compared to the text-based test set indicates that the radiology reports themselves might not encapsulate all the data needed to accurately classify the presence or absence of a pathology, and that labels generated synthetically from radiology reports might not be reliable for training image pathology classifiers using supervised learning . We encourage research aiming at improving the synthetic generation of pathology labels based on text. 

\subsection{RadGraph Annotations}
\label{sec:radgraph}

\begin{table}[ht!]
\centering
\caption{RadGraph annotations generated for the Findings and Impression section of the reports.}
\label{table:radgraph}
\resizebox{0.5\textwidth}{!}{%
\begin{tabular}{@{}lrr@{}}
\toprule[1.5pt]
\cellcolor[HTML]{EFEFEF}Category & \cellcolor[HTML]{EFEFEF}Findings (\%) & \cellcolor[HTML]{EFEFEF}Impression (\%)\\
\midrule
Anatomy & 740,453 (46.81) & 1,730,617 (43.27) \\
Observation Present & 685,006 (43.30) & 1,762,298 (44.06) \\
Observation Uncertain & 58,772 (3.72) & 206,959 (5.17) \\
Observation Absent & 97,234 (6.15) & 298,811 (7.47) \\
\midrule
Total Entities & 1,581,863 (100) & 3,999,559 (100) \\
\midrule
Modify & 658,095 (59.47) & 1,640,192 (59.16) \\
Located at & 400,132 (36.16) & 977,785 (35.27) \\
Suggestive of & 48,432 (4.38) & 154,360 (5.57) \\
\midrule
Total Relations & 1,106,659 (100) & 2,772,337 (100) \\
\bottomrule[1.5pt]
\end{tabular}
}
\end{table}
The Table~\ref{table:radgraph} showcases the RadGraph \cite{jain2021radgraph} annotations released as part of \name~for the Findings and Impression sections of the radiology reports. The most common annotations are Anatomy (46.81\% in Findings and 43.27\% in Impression) and Observation: Definitely Present (43.30\% in Findings and 44.06\% in Impression). Observation: Uncertain and Definitely Absent categories are less frequent, ranging from 3.72\% to 7.47\%. Relation annotations like Modify (59.47\% in Findings and 59.16\% in Impression) dominate over Located at and Suggestive of. The total entity annotations are significant, with 1,581,863 in Findings and 3,999,559 in Impression, while total relations annotations are 1,106,659 (Findings) and 2,772,337 (Impression), demonstrating the comprehensive nature of the dataset.

The Table~\ref{table:radgraph} showcases the generated RadGraph annotations~\cite{jain2021radgraph} on our dataset. These annotations cover the Findings and Impression sections of radiology reports. The most frequent annotations are "Anatomy" (46.81\% in Findings and 43.27\% in Impression) and "Observation: Definitely Present" (43.30\% in Findings and 44.06\% in Impression). The categories "Observation: Uncertain" and "Definitely Absent" are less common, with frequencies ranging from 3.72\% to 7.47\%. Relation annotations, such as "Modify" (59.47\% in Findings and 59.16\% in Impression), are more prevalent than "Located at" and "Suggestive of." Entity annotations amount to 1,581,863 in Findings and 3,999,559 in Impression, while the total number of relation annotations are 1,106,659 (in Findings) and 2,772,337 (in Impression).
\subsection{Model Zoo} 
\label{sec:modelzoo}
As part of \name~, we are releasing pretrained models  trained on \name~data, incorporating recent developments in machine learning spanning natural language processing, image recognition, and generative modeling. Among these releases are a pretrained LLaMA~\cite{touvron2023llama} model, which generates human-like text and excels at complex language tasks. A pretrained CLIP~\cite{radford2021learning} model is being introduced, which improves the way visual concepts are learned from text descriptions, thereby enhancing image search and classification capabilities. The lineup also includes a pretrained VQ-GAN~\cite{esser2021taming} model, blending VQ-VAE and GAN technologies to produce realistic images and demonstrating generative power. The release also includes a pretrained DINOv2~\cite{oquab2023dinov2} model that employs self-supervised learning with Vision Transformers to achieve robust visual representations. Finally, several architecture-agnostic models are included, todeliver competitive performance in radiology report generation (RRG) and radiology report summarization (RRS).
\\
These models are available at the following URL:~\url{https://github.com/Stanford-AIMI/chexpert-plus}.
\section{Dataset Release and Analysis}

In this section, we will explore how the release of \name~dataset was made possible, and analyze uses and possible future work.

We begin with a comparison between \name~ and the original CheXpert 1.0 dataset in Section~\ref{sec:comparison_1.0}, as well as other existing datasets in Section~\ref{sec:comparison_others}. Then, we take an in-depth look at the de-identification procedures applied to both the radiology images and reports, which are explained in Section~\ref{sec:dedim} and Section~\ref{sec:deidreport}, respectively. Finally, we discuss the applicability of this dataset in Section~\ref{section:applicability} and the limitations in Section~\ref{section:limitations}.

\subsection{Comparison with CheXpert 1.0}
\label{sec:comparison_1.0}

Building upon the foundational CheXpert dataset introduced in 2019, which comprised 224,316 chest radiographs from 65,240 patients annotated for 14 observations, our enhanced version, \name, introduces several significant advancements aimed at pushing the boundaries of medical imaging research. By transitioning to DICOM, the gold standard in medical image formats, we provide images alongside a subset of their original DICOM headers, thereby offering richer image metadata and superior image quality. \name~further enriches the dataset by including the corresponding radiology reports, which we have parsed into sections such as medical history, findings, and impressions, enhancing the dataset's utility for comprehensive analysis. Moreover, by incorporating detailed patient demographic data, \name~facilitates the development of fairness-focused analyses and multimodal models capable of utilizing this information for more informed and nuanced diagnoses.

In addition, \name~also focuses on releasing higher quality extracted labels, be it for the 14 lung diseases thanks to the CheXbert annotation tool, or for the more recent RadGraph annotations, allowing the training of classifiers directly on top of \name~or for CheXpert- and RadGraph-based metrics to be computed.

These enhancements not only deepen the analytical potential within the medical imaging field but also significantly contribute to the evolving medical AI community, particularly as it strides towards leveraging multimodal and image-text learning paradigms, promising to enhance diagnostic accuracy and patient outcomes.

Finally, we mention that 186 studies are missing compared to the original CheXpert release, as the corresponding reports could not be recovered when preparing the \name~release.
\subsection{Head-to-head Comparison with other Radiology Datasets}
\label{sec:comparison_others}

While numerous chest X-ray datasets exist (Table~\ref{tab:related_dataset}), few compare \name~in terms of scale and modality inclusiveness. Among these, MIMIC, OpenI, and PadChest also provide radiology reports, yet PadChest's utilization of Spanish and OpenI's smaller scale differentiate \name~significantly. Additionally, besides MIMIC and OpenI, only CandidPTX, and VinDR offer X-rays in DICOM format and neither matches \name~in the sheer volume of studies.

MIMIC, the closest equivalent, features 377,095 chest X-rays with accompanying reports and includes both DICOM images and patient demographic information. \name~ surpasses MIMIC in textual depth, containing 36 million text tokens compared to MIMIC's 34 million. MIMIC does features a higher number of reports (227,821 unique reports compared to \name's 187,711 unique reports), these reports are shorter, especially in their impression section: MIMIC impressions count 7,986,317 total tokens, versus the 13,351,758 tokens of \name's impressions. For any text-image radiology task where the text being handled is restrained to the impression section, \name~is therefore also the largest dataset available. Finally, our reports underwent a de-identification process that did not alter their structure, even including de-identified accession number sections, something that is not displayed in MIMIC and makes our reports closer to the natural textual data handled during radiology exams.

Given these distinctions and advancements, \name~represents a valuable addition to the existing landscape of medical imaging datasets, poised to significantly enhance and expand the capabilities of research in medical AI.

\subsection{De-identification and release of DICOMs}
\label{sec:dedim}

The release of DICOMs required the de-identification of all metadata DICOM headers, as well as their pixel content, as displayed in Figure~\ref{fig:figure_chexpert_deid}. The metadata headers were automatically de-identified and reviewed by humans to confirm that no PHI information was present (corresponding to steps 5 and 6 of Figure~\ref{fig:figure_chexpert_deid}). Along that, we guarantee that the pixels in the released DICOMs are identical to those in the original CheXpert 1.0 images, that were themselves cleared for public release by hiding any PHI content (steps 7 and 8). The code used to accomplish this is available in Appendix~\ref{app:code_dicom}.

\subsection{De-identification and release of reports}
\label{sec:deidreport}

\begin{figure*}[!ht]
\includegraphics[width=1\textwidth]{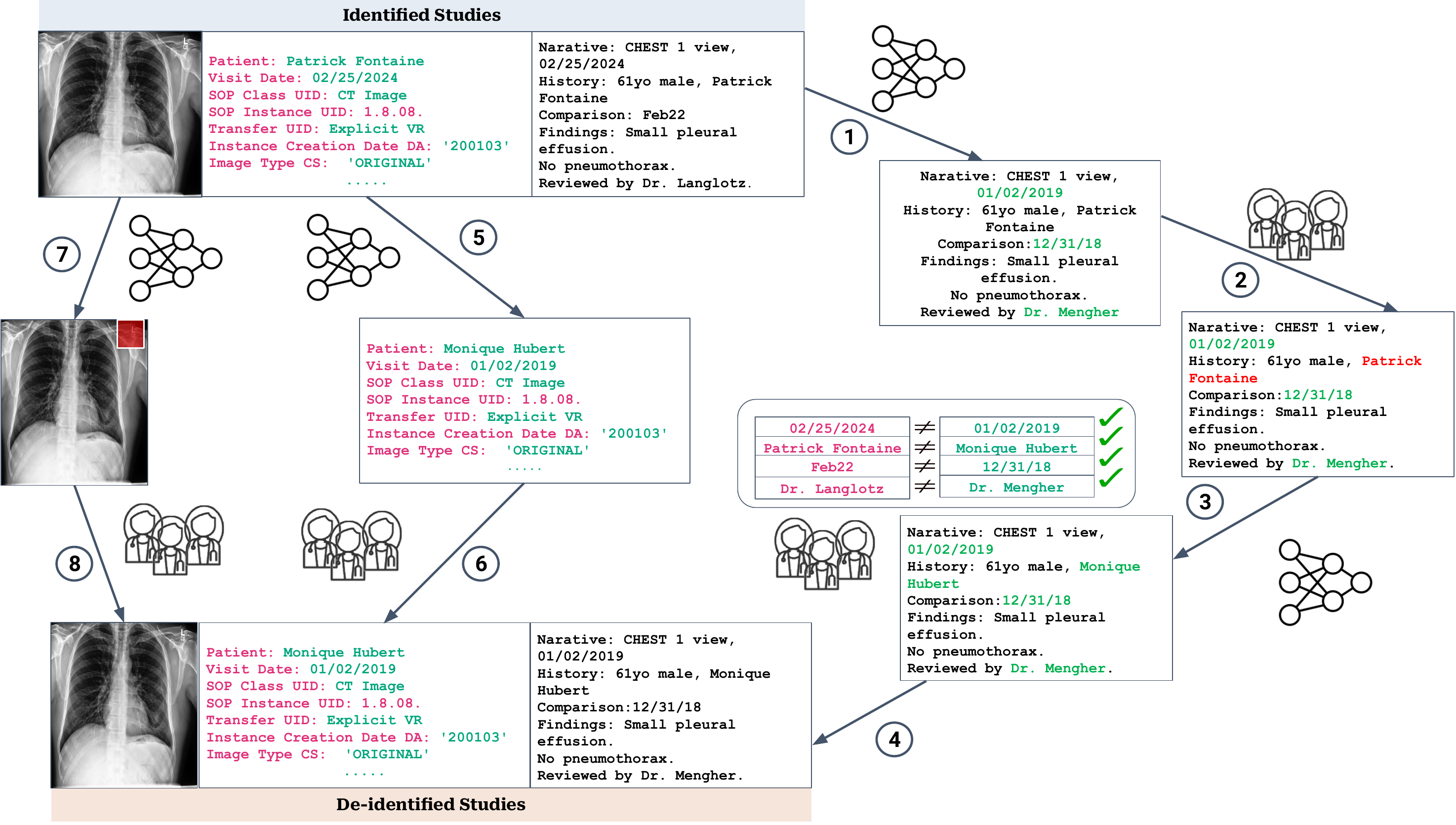}
\caption{The de-identification of images and reports from \name~is an 8-step process as described in Section~\ref{sec:deidreport}.}
\label{fig:figure_chexpert_deid}
\end{figure*}

As part of the \name~release, the reports associated with the CheXpert images underwent a  de-identification process that lasted for a year and was supported by 25-30 human annotators. We counted a posteriori the presence of 853,878 total PHI spans, as defined in the Health Insurance Portability and Accountability Act of 1996 (HIPAA). Details of the types of PHI spotted in these reports are further displayed in Table~\ref{table:data_summary}.

\begin{table}[!h]
\centering
\begin{tabular}{@{}lr@{}}
\toprule[1.5pt]
\multicolumn{1}{c}{\cellcolor[HTML]{EFEFEF}Data Category} & \multicolumn{1}{c}{\cellcolor[HTML]{EFEFEF}Count} \\ \midrule
Dates                                                     & 538,160                                           \\
Age numbers
& 5,206                                             \\
Unique identifiers                                        & 233,244                                           \\
Healthcare worker names                                   & 57,274                                            \\
Vendor names                                              & 9,110                                             \\
Phone numbers                                             & 10,324                                            \\
Hospital names                                            & 443                                               \\
Patient names                                             & 10,324                                            \\ \bottomrule[1.5pt]
\end{tabular}
\caption{Data Summary}
\label{table:data_summary}
\end{table}

To the best of our knowledge, this is the largest text de-identification effort in terms of quantity of PHI reviewed.
As displayed in Figure~\ref{fig:figure_chexpert_deid}, these reports underwent a process in 4 steps in order to be considered de-identified and ready for public release.

\textbf{Step 1} The reports were first automatically de-identified using a two-step model ~\cite{10.1093/jamia/ocac219} that first leverages a transformer model for token-level classification into PHI categories, before replacing the true PHI spans with synthetic PHI as a "Hide in plain sight" approach ~\cite{Carrell_Malin_Aberdeen_Bayer_Clark_Wellner_Hirschman_2013a}. This latter approach adds an additional safety factor to ensure any missed PHI, if any, cannot be easily spotted.

\textbf{Step 2} Human annotators reviewed each report with the synthetic PHI highlighted by the automatic de-identifier The human reviewers identified any PHI that was missed by the algorithm. A missed PHI can be a full span belonging to a certain PHI category, or any prefix or suffix of a PHI span partially detected by the model. Any missed PHI was  highlighted by the annotators for further human review. Out of the 853,878 true PHI spans in all reports, 23 were fully missed by the model (0.002\%), and 841 were partially missed by the model (0.01\%).

\textbf{Step 3} Any partially or fully missed PHI then underwent the same "Hide In Plain Sight" step to replace the remaining true PHI spans  by synthetic PHI spans.

\textbf{Step 4} A mapping of each pair of true and synthetic PHI spans was generated and reviewed for PHI by a board-certified radiologist.  This combination of automated processes and human review confirmed that all true PHI spans had been replaced by synthetic PHI spans. 

\subsection{Applicability, Usage and Future Work}
\label{section:applicability}

\name~dataset is being released on the Stanford AIMI Shared Dataset website and can be accessed at the following link. It is associated with a Stanford University Dataset Research Use Agreement, which specifies that \name~may not be used for any commercial purposes and is only available for research uses. In particular, you may not distribute, publish or reproduce a copy of this dataset.

As part of the \name~release, we underline the following main uses that can be made out of it:

\textbf{Performance}: Due to its extensive size, \name~almost double the amount of publicly available English text-image pairs in radiology. Therefore, any model can leverage this dataset to improve in performance compared to models trained before this data release.

\textbf{Robustness}: Along MIMIC, \name~is the only large size english text-image dataset in radiology, therefore allowing to not only test models on multiple institutions but also perform cross-institution training, hopefully leading to more robust performance when evaluating on completely new institutions.

\textbf{Fairness}: The inclusion of extensive amounts of patient basic demographic data in \name~enables downstream applications to account for the imbalance of patient ethnicity, sex, age or socio-economic background, therefore potentially limiting the bias of models trained for various radiology tasks.

Finally, we underline that any model trained with the help of \name~data may still reflect biases based on patient characteristics and pathologies, among else. When using such models, researchers should always look for sources of potential distribution shifts and audit for peformance disparities based on attributes such as race, ethnicity, age or socio-economic background.

\subsection{Limitations of the dataset}
\label{section:limitations}

First, as described in Section~\ref{sec:reports}, the focus was put on collecting reports with detailed impression sections, at the cost of having lots of findings. Therefore, impressions are in total of tokens twice as important as findings. Second, some pathologies such as fracture, lung lesion and pleural other are significantly under-represented, as displayed in Section~\ref{sec:chexpertlabels}, which limits the performance of any models aiming at studying these particular pathologies. 
\section{Conclusion}

In our work, we introduced \name~, a multi-sourced dataset comprising hundreds of thousands of images paired with texts, patient demographics, and computed pathology labels and RadGraph annotations. First, we release high-quality images in DICOM format along with DICOM metadata encapsulating image processing information, as well as PNG images. Second, we distribute the corresponding reports after a careful de-identification process and pre-parse these reports into their corresponding subsections for ease of use in various downstream tasks. Third, we provide patient metadata detailing clinical and socio-economic conditions such as sex, ethnicity, and medical insurance information, for better analysis of distribution shifts and the diminishment of biases. Fourth, we release improved pathology labels to be used directly for classification or evaluation tasks. Fifth, we pre-compute RadGraph annotations for both findings and impressions, making them available to be directly usable in existing pipelines. Finally, we release a set of models for the main radiology downstream tasks, spanning from text-to-image generation to text-to-text summarization. We hope this substantial data release helps foster the development of AI models in radiology that display improved performance, robustness, and fairness and ultimately improve patient medical care.

\section{Acknowledgements}

This work was supported in part by MIDRC (The Medical Imaging and Data Resource Center), funded by the National Institute of Biomedical Imaging and Bioengineering (NIBIB) of the National Institutes of Health under contract 75N92020D00021.
\\
This work has been made possible thanks to the help of Stephanie Bogdan and her contributions to gather the dataset, before proceeding to its de-identification and cleaning.
\\
We would also like to thank Bui Duc Thai Tan and Duong Thi Hong Hanh for all their help in the de-identification and the release of the \name~reports.

\bibliography{anthology,custom}

\begin{thebibliography}{33}
\expandafter\ifx\csname natexlab\endcsname\relax\def\natexlab#1{#1}\fi

\bibitem[{Bannur et~al.(2023)Bannur, Hyland, Liu, Perez-Garcia, Ilse, Castro, Boecking, Sharma, Bouzid, Thieme et~al.}]{bannur2023learning}
Shruthi Bannur, Stephanie Hyland, Qianchu Liu, Fernando Perez-Garcia, Maximilian Ilse, Daniel~C Castro, Benedikt Boecking, Harshita Sharma, Kenza Bouzid, Anja Thieme, et~al. 2023.
\newblock Learning to exploit temporal structure for biomedical vision-language processing.
\newblock In \emph{Proceedings of the IEEE/CVF Conference on Computer Vision and Pattern Recognition}, pages 15016--15027.

\bibitem[{Bustos et~al.(2020)Bustos, Pertusa, Salinas, and De~La Iglesia-Vaya}]{bustos2020padchest}
Aurelia Bustos, Antonio Pertusa, Jose-Maria Salinas, and Maria De~La Iglesia-Vaya. 2020.
\newblock Padchest: A large chest x-ray image dataset with multi-label annotated reports.
\newblock \emph{Medical image analysis}, 66:101797.

\bibitem[{Carrell et~al.(2013)Carrell, Malin, Aberdeen, Bayer, Clark, Wellner, and Hirschman}]{Carrell_Malin_Aberdeen_Bayer_Clark_Wellner_Hirschman_2013a}
David Carrell, Bradley Malin, John Aberdeen, Samuel Bayer, Cheryl Clark, Ben Wellner, and Lynette Hirschman. 2013.
\newblock \href {https://www.ncbi.nlm.nih.gov/pmc/articles/PMC3638183/} {Hiding in plain sight: Use of realistic surrogates to reduce exposure of protected health information in clinical text}.

\bibitem[{Chambon et~al.(2022{\natexlab{a}})Chambon, Bluethgen, Delbrouck, Van~der Sluijs, Po{\l}acin, Chaves, Abraham, Purohit, Langlotz, and Chaudhari}]{chambon2022roentgen}
Pierre Chambon, Christian Bluethgen, Jean-Benoit Delbrouck, Rogier Van~der Sluijs, Ma{\l}gorzata Po{\l}acin, Juan Manuel~Zambrano Chaves, Tanishq~Mathew Abraham, Shivanshu Purohit, Curtis~P Langlotz, and Akshay Chaudhari. 2022{\natexlab{a}}.
\newblock Roentgen: Vision-language foundation model for chest x-ray generation.
\newblock \emph{arXiv preprint arXiv:2211.12737}.

\bibitem[{Chambon et~al.(2022{\natexlab{b}})Chambon, Wu, Steinkamp, Adleberg, Cook, and Langlotz}]{10.1093/jamia/ocac219}
Pierre~J Chambon, Christopher Wu, Jackson~M Steinkamp, Jason Adleberg, Tessa~S Cook, and Curtis~P Langlotz. 2022{\natexlab{b}}.
\newblock \href {https://doi.org/10.1093/jamia/ocac219} {{Automated deidentification of radiology reports combining transformer and “hide in plain sight” rule-based methods}}.
\newblock \emph{Journal of the American Medical Informatics Association}, 30(2):318--328.

\bibitem[{Chaves et~al.(2024)Chaves, Huang, Xu, Xu, Usuyama, Zhang, Wang, Xie, Khademi, Yang et~al.}]{chaves2024training}
Juan Manuel~Zambrano Chaves, Shih-Cheng Huang, Yanbo Xu, Hanwen Xu, Naoto Usuyama, Sheng Zhang, Fei Wang, Yujia Xie, Mahmoud Khademi, Ziyi Yang, et~al. 2024.
\newblock Training small multimodal models to bridge biomedical competency gap: A case study in radiology imaging.
\newblock \emph{arXiv preprint arXiv:2403.08002}.

\bibitem[{Chen et~al.(2020)Chen, Song, Chang, and Wan}]{chen2020generating}
Zhihong Chen, Yan Song, Tsung-Hui Chang, and Xiang Wan. 2020.
\newblock Generating radiology reports via memory-driven transformer.
\newblock In \emph{Proceedings of the 2020 Conference on Empirical Methods in Natural Language Processing (EMNLP)}, pages 1439--1449.

\bibitem[{Chen et~al.(2024)Chen, Varma, Delbrouck, Paschali, Blankemeier, Van~Veen, Valanarasu, Youssef, Cohen, Reis et~al.}]{chen2024chexagent}
Zhihong Chen, Maya Varma, Jean-Benoit Delbrouck, Magdalini Paschali, Louis Blankemeier, Dave Van~Veen, Jeya Maria~Jose Valanarasu, Alaa Youssef, Joseph~Paul Cohen, Eduardo~Pontes Reis, et~al. 2024.
\newblock Chexagent: Towards a foundation model for chest x-ray interpretation.
\newblock \emph{arXiv preprint arXiv:2401.12208}.

\bibitem[{Delbrouck et~al.(2022)Delbrouck, Chambon, Bluethgen, Tsai, Almusa, and Langlotz}]{delbrouck2022improving}
Jean-Benoit Delbrouck, Pierre Chambon, Christian Bluethgen, Emily Tsai, Omar Almusa, and Curtis Langlotz. 2022.
\newblock Improving the factual correctness of radiology report generation with semantic rewards.
\newblock In \emph{Findings of the Association for Computational Linguistics: EMNLP 2022}, pages 4348--4360.

\bibitem[{Demner-Fushman et~al.(2012)Demner-Fushman, Antani, Simpson, and Thoma}]{demner2012design}
Dina Demner-Fushman, Sameer Antani, Matthew Simpson, and George~R Thoma. 2012.
\newblock Design and development of a multimodal biomedical information retrieval system.
\newblock \emph{Journal of Computing Science and Engineering}, 6(2):168--177.

\bibitem[{Esser et~al.(2021)Esser, Rombach, and Ommer}]{esser2021taming}
Patrick Esser, Robin Rombach, and Bjorn Ommer. 2021.
\newblock Taming transformers for high-resolution image synthesis.
\newblock In \emph{Proceedings of the IEEE/CVF conference on computer vision and pattern recognition}, pages 12873--12883.

\bibitem[{Feng et~al.(2021{\natexlab{a}})Feng, Azzollini, Kim, Jin, Gordon, Yeoh, Kim, Han, Lee, Patel et~al.}]{feng2021curation}
Sijing Feng, Damian Azzollini, Ji~Soo Kim, Cheng-Kai Jin, Simon~P Gordon, Jason Yeoh, Eve Kim, Mina Han, Andrew Lee, Aakash Patel, et~al. 2021{\natexlab{a}}.
\newblock Curation of the candid-ptx dataset with free-text reports.
\newblock \emph{Radiology: Artificial Intelligence}, 3(6):e210136.

\bibitem[{Feng et~al.(2021{\natexlab{b}})Feng, Azzollini, Kim, Jin, Kim, Gordon, Yeoh, Han, Lee, Patel, Urschler, Fong, Simmers, Tarr, Barnard, and Wilson}]{Feng2021}
Sijing Feng, Damian Azzollini, Ji~Soo Kim, Cheng~Kai Jin, Eve Kim, Simon Gordon, Jason Yeoh, Min~A Han, Andrew Lee, Aakash Patel, Martin Urschler, Amy Fong, Cameron Simmers, Gregory Tarr, Stuart Barnard, and Ben Wilson. 2021{\natexlab{b}}.
\newblock \href {https://doi.org/10.17608/k6.auckland.14173982} {{CANDID-PTX}}.
\newblock \emph{Radiology: Artificial Intelligence}.

\bibitem[{Healthcare()}]{jfhealthcare}
JF~Healthcare.
\newblock Object-cxr - automatic detection of foreign objects on chest x-rays.
\newblock \url{https://web.archive.org/web/20201127235812/https://jfhealthcare.github.io/object-CXR/}.

\bibitem[{Huang et~al.(2023)Huang, Pareek, Jensen, Lungren, Yeung, and Chaudhari}]{huang2023self}
Shih-Cheng Huang, Anuj Pareek, Malte Jensen, Matthew~P Lungren, Serena Yeung, and Akshay~S Chaudhari. 2023.
\newblock Self-supervised learning for medical image classification: a systematic review and implementation guidelines.
\newblock \emph{NPJ Digital Medicine}, 6(1):74.

\bibitem[{Huang et~al.(2021)Huang, Shen, Lungren, and Yeung}]{huang2021gloria}
Shih-Cheng Huang, Liyue Shen, Matthew~P Lungren, and Serena Yeung. 2021.
\newblock Gloria: A multimodal global-local representation learning framework for label-efficient medical image recognition.
\newblock In \emph{Proceedings of the IEEE/CVF International Conference on Computer Vision}, pages 3942--3951.

\bibitem[{Hyland et~al.(2023)Hyland, Bannur, Bouzid, Castro, Ranjit, Schwaighofer, P{\'e}rez-Garc{\'\i}a, Salvatelli, Srivastav, Thieme et~al.}]{hyland2023maira}
Stephanie~L Hyland, Shruthi Bannur, Kenza Bouzid, Daniel~C Castro, Mercy Ranjit, Anton Schwaighofer, Fernando P{\'e}rez-Garc{\'\i}a, Valentina Salvatelli, Shaury Srivastav, Anja Thieme, et~al. 2023.
\newblock Maira-1: A specialised large multimodal model for radiology report generation.
\newblock \emph{arXiv preprint arXiv:2311.13668}.

\bibitem[{Irvin et~al.(2019)Irvin, Rajpurkar, Ko, Yu, Ciurea-Ilcus, Chute, Marklund, Haghgoo, Ball, Shpanskaya et~al.}]{irvin2019chexpert}
Jeremy Irvin, Pranav Rajpurkar, Michael Ko, Yifan Yu, Silviana Ciurea-Ilcus, Chris Chute, Henrik Marklund, Behzad Haghgoo, Robyn Ball, Katie Shpanskaya, et~al. 2019.
\newblock Chexpert: A large chest radiograph dataset with uncertainty labels and expert comparison.
\newblock In \emph{Proceedings of the AAAI conference on artificial intelligence}, volume~33, pages 590--597.

\bibitem[{Jain et~al.(2021)Jain, Agrawal, Saporta, Truong, Duong, Bui, Chambon, Zhang, Lungren, Ng, Langlotz, and Rajpurkar}]{jain2021radgraph}
Saahil Jain, Ashwin Agrawal, Adriel Saporta, Steven~QH Truong, Du~Nguyen Duong, Tan Bui, Pierre Chambon, Yuhao Zhang, Matthew~P. Lungren, Andrew~Y. Ng, Curtis~P. Langlotz, and Pranav Rajpurkar. 2021.
\newblock \href {http://arxiv.org/abs/2106.14463} {Radgraph: Extracting clinical entities and relations from radiology reports}.

\bibitem[{Johnson et~al.(2019)Johnson, Pollard, Berkowitz, Greenbaum, Lungren, Deng, Mark, and Horng}]{johnson2019mimic}
Alistair~EW Johnson, Tom~J Pollard, Seth~J Berkowitz, Nathaniel~R Greenbaum, Matthew~P Lungren, Chih-ying Deng, Roger~G Mark, and Steven Horng. 2019.
\newblock Mimic-cxr, a de-identified publicly available database of chest radiographs with free-text reports.
\newblock \emph{Scientific data}, 6(1):317.

\bibitem[{Moor et~al.(2023)Moor, Huang, Wu, Yasunaga, Dalmia, Leskovec, Zakka, Reis, and Rajpurkar}]{moor2023med}
Michael Moor, Qian Huang, Shirley Wu, Michihiro Yasunaga, Yash Dalmia, Jure Leskovec, Cyril Zakka, Eduardo~Pontes Reis, and Pranav Rajpurkar. 2023.
\newblock Med-flamingo: a multimodal medical few-shot learner.
\newblock In \emph{Machine Learning for Health (ML4H)}, pages 353--367. PMLR.

\bibitem[{Nguyen et~al.(2020)Nguyen, Pham, Nguyen, Nguyen, Dao, Vu, Lam, and Le}]{nguyen2020vinbigdata}
H~Nguyen, HH~Pham, NT~Nguyen, DB~Nguyen, M~Dao, V~Vu, K~Lam, and LT~Le. 2020.
\newblock Vinbigdata chest x-ray abnormalities detection.
\newblock \emph{Kaggle Competition https://www. kaggle. com/c/vinbi gdatachest-xray-abnor malit ies-detec tion}.

\bibitem[{Oquab et~al.(2023)Oquab, Darcet, Moutakanni, Vo, Szafraniec, Khalidov, Fernandez, Haziza, Massa, El-Nouby et~al.}]{oquab2023dinov2}
Maxime Oquab, Timoth{\'e}e Darcet, Th{\'e}o Moutakanni, Huy Vo, Marc Szafraniec, Vasil Khalidov, Pierre Fernandez, Daniel Haziza, Francisco Massa, Alaaeldin El-Nouby, et~al. 2023.
\newblock Dinov2: Learning robust visual features without supervision.
\newblock \emph{arXiv preprint arXiv:2304.07193}.

\bibitem[{Radford et~al.(2021)Radford, Kim, Hallacy, Ramesh, Goh, Agarwal, Sastry, Askell, Mishkin, Clark et~al.}]{radford2021learning}
Alec Radford, Jong~Wook Kim, Chris Hallacy, Aditya Ramesh, Gabriel Goh, Sandhini Agarwal, Girish Sastry, Amanda Askell, Pamela Mishkin, Jack Clark, et~al. 2021.
\newblock Learning transferable visual models from natural language supervision.
\newblock In \emph{International conference on machine learning}, pages 8748--8763. PMLR.

\bibitem[{Reis et~al.(2022)Reis, de~Paiva, da~Silva, Ribeiro, Paiva, Bulgarelli, Lee, Santos, Brito, Amaral et~al.}]{reis2022brax}
Eduardo~P Reis, Joselisa~PQ de~Paiva, Maria~CB da~Silva, Guilherme~AS Ribeiro, Victor~F Paiva, Lucas Bulgarelli, Henrique~MH Lee, Paulo~V Santos, Vanessa~M Brito, Lucas~TW Amaral, et~al. 2022.
\newblock Brax, brazilian labeled chest x-ray dataset.
\newblock \emph{Scientific Data}, 9(1):487.

\bibitem[{Smit et~al.(2020)Smit, Jain, Rajpurkar, Pareek, Ng, and Lungren}]{smit2020chexbert}
Akshay Smit, Saahil Jain, Pranav Rajpurkar, Anuj Pareek, Andrew~Y. Ng, and Matthew~P. Lungren. 2020.
\newblock \href {http://arxiv.org/abs/2004.09167} {Chexbert: Combining automatic labelers and expert annotations for accurate radiology report labeling using bert}.

\bibitem[{Touvron et~al.(2023)Touvron, Martin, Stone, Albert, Almahairi, Babaei, Bashlykov, Batra, Bhargava, Bhosale et~al.}]{touvron2023llama}
Hugo Touvron, Louis Martin, Kevin Stone, Peter Albert, Amjad Almahairi, Yasmine Babaei, Nikolay Bashlykov, Soumya Batra, Prajjwal Bhargava, Shruti Bhosale, et~al. 2023.
\newblock Llama 2: Open foundation and fine-tuned chat models.
\newblock \emph{arXiv preprint arXiv:2307.09288}.

\bibitem[{Tu et~al.(2024)Tu, Azizi, Driess, Schaekermann, Amin, Chang, Carroll, Lau, Tanno, Ktena et~al.}]{tu2024towards}
Tao Tu, Shekoofeh Azizi, Danny Driess, Mike Schaekermann, Mohamed Amin, Pi-Chuan Chang, Andrew Carroll, Charles Lau, Ryutaro Tanno, Ira Ktena, et~al. 2024.
\newblock Towards generalist biomedical ai.
\newblock \emph{NEJM AI}, 1(3):AIoa2300138.

\bibitem[{Varma et~al.(2023)Varma, Delbrouck, Hooper, Chaudhari, and Langlotz}]{varma2023villa}
Maya Varma, Jean-Benoit Delbrouck, Sarah Hooper, Akshay Chaudhari, and Curtis Langlotz. 2023.
\newblock Villa: Fine-grained vision-language representation learning from real-world data.
\newblock In \emph{Proceedings of the IEEE/CVF International Conference on Computer Vision}.

\bibitem[{Vay{\'a} et~al.(2020)Vay{\'a}, Saborit, Montell, Pertusa, Bustos, Cazorla, Galant, Barber, Orozco-Beltr{\'a}n, Garc{\'\i}a-Garc{\'\i}a et~al.}]{vaya2020bimcv}
Maria De La~Iglesia Vay{\'a}, Jose~Manuel Saborit, Joaquim~Angel Montell, Antonio Pertusa, Aurelia Bustos, Miguel Cazorla, Joaquin Galant, Xavier Barber, Domingo Orozco-Beltr{\'a}n, Francisco Garc{\'\i}a-Garc{\'\i}a, et~al. 2020.
\newblock Bimcv covid-19+: a large annotated dataset of rx and ct images from covid-19 patients.
\newblock \emph{arXiv preprint arXiv:2006.01174}.

\bibitem[{Wang et~al.(2017)Wang, Peng, Lu, Lu, Bagheri, and Summers}]{wang2017chestx}
Xiaosong Wang, Yifan Peng, Le~Lu, Zhiyong Lu, Mohammadhadi Bagheri, and Ronald~M Summers. 2017.
\newblock Chestx-ray8: Hospital-scale chest x-ray database and benchmarks on weakly-supervised classification and localization of common thorax diseases.
\newblock In \emph{Proceedings of the IEEE conference on computer vision and pattern recognition}, pages 2097--2106.

\bibitem[{Wu et~al.(2023)Wu, Zhang, Zhang, Wang, and Xie}]{wu2023towards}
Chaoyi Wu, Xiaoman Zhang, Ya~Zhang, Yanfeng Wang, and Weidi Xie. 2023.
\newblock Towards generalist foundation model for radiology.
\newblock \emph{arXiv preprint arXiv:2308.02463}.

\bibitem[{Zhang et~al.(2022)Zhang, Jiang, Miura, Manning, and Langlotz}]{zhang2022contrastive}
Yuhao Zhang, Hang Jiang, Yasuhide Miura, Christopher~D Manning, and Curtis~P Langlotz. 2022.
\newblock Contrastive learning of medical visual representations from paired images and text.
\newblock In \emph{Machine Learning for Healthcare Conference}, pages 2--25. PMLR.

\end{thebibliography}
\bibliographystyle{acl_natbib}

\clearpage
\appendix

\onecolumn

\section{From \name~DICOM to CheXpert 1.0 JPG} \label{app:code_dicom}

\textbf{Listing 1:} Python code to convert \name~DICOM to CheXpert 1.0 JPG.
\begin{lstlisting}[language=Python]
import pydicom
import cv2
import os

def convert_dicom_to_images(input_file_path, jpg_filename):
    """
    This function converts a DICOM file to a JPEG image.

    Args:
        input_file_path (str): The path to the input DICOM file.
        jpg_filename (str): The filename (including the path) to save the JPEG image.

    Returns:
        None
    """
    # Read the DICOM file
    dcm_file = pydicom.dcmread(input_file_path)

    # Rescale the pixel array to the range [0, 255]
    rescaled_image = cv2.convertScaleAbs(dcm_file.pixel_array, alpha=(255.0 / dcm_file.pixel_array.max()))

    # If the PhotometricInterpretation is "MONOCHROME1", invert the pixel values
    if dcm_file.PhotometricInterpretation == "MONOCHROME1":
        rescaled_image = cv2.bitwise_not(rescaled_image)

    # Apply histogram equalization to enhance the contrast
    adjusted_image = cv2.equalizeHist(rescaled_image)

    # Save the adjusted image in JPG format using the specified output file path
    cv2.imwrite(jpg_filename, adjusted_image)


# Define your paths and filenames
dcm_path = 'path/to/filename.dcm'
jpg_path = 'path/to/filename.jpg'

# Check if the DICOM file exists
if os.path.isfile(dcm_path):
    # Convert DICOM to JPEG and compare
    convert_dicom_to_images(dcm_path, jpg_path)
else:
    print(f"The DICOM file '{dcm_path}' does not exist.")
\end{lstlisting}
\twocolumn 

\section{DICOM metadata}\label{app:dicom_meta}

\renewcommand{\arraystretch}{0.95}
\begin{table}[h!]
\begin{tabular}{@{}llc@{}}
\toprule[1.5pt]
\multicolumn{1}{c}{\cellcolor[HTML]{EFEFEF}DICOM metadata}                          \\ \midrule
                                  PixelData
    \\ BitsAllocated
    \\ Rows
    \\ Columns
    \\ SamplesPerPixel
    \\ PhotometricInterpretation 
    \\ PixelRepresentation
    \\ BitsStored
    \\ ImagePositionPatient
    \\ PixelSpacing
    \\ RescaleIntercept
    \\ RescaleSlope
    \\ WindowCenter
    \\ WindowWidth
    \\ Manufacturer
    \\ SliceThickness
    \\ ImageOrientationPatient
    \\ VOILUTFunction
    \\ VOILUTSequence
    \\ PresentationLUTShape
    \\ LUTExplanation
    \\ Exposure
    \\ ExposureControlMode
    \\ ExposureControlModeDescription
    \\ ExposureInuAs
    \\ RelativeXRayExposure
    \\ ExposuresOnPlate
    \\ ExposureIndex
    \\ TargetExposureIndex
    \\ ExposureTimeInuS
    \\ ExposuresOnDetectorSinceLastCalibration
    \\ DetectorTimeSinceLastExposure
    \\ TotalNumberOfExposures
    \\ ExposureStatus
    \\ ExposureTime
    \\ ExposureInmAs
    \\ ExposureModulationType
    \\ KVP
    \\ Laterality
    \\ ImageLaterality
    \\ RescaleType
    \\ XRayTubeCurrent
    \\ XRayTubeCurrentInuA
    \\ ConvolutionKernel
    \\ ViewPosition
    \\ BodyPartExamined
    \\ BurnedInAnnotation           
                  \\  \bottomrule[1.5pt]
\end{tabular}
\caption{Combination of metadata contained in \name~DICOM files.}
\label{table:dicom_metadata}
\end{table}
\end{document}